\documentclass[10pt,conference]{IEEEtran}
\usepackage{graphicx}
\usepackage{booktabs}
\usepackage{amsmath}
\usepackage{amssymb}
\usepackage{xcolor}
\usepackage{multirow}
\usepackage{array}
\usepackage{cite}
\usepackage{url}
\usepackage{hyperref}
\usepackage{algorithm}
\usepackage{algorithmic}
\usepackage{subfigure}
\usepackage{balance}

\hypersetup{
    colorlinks=true,
    linkcolor=blue,
    filecolor=magenta,      
    urlcolor=cyan,
    citecolor=green,
    pdftitle={Is GPT-OSS All You Need? Benchmarking Large Language Models for Financial Intelligence and the Surprising Efficiency Paradox},
    pdfauthor={Anonymous Authors},
}

\begin{document}

\title{Is GPT-OSS All You Need? Benchmarking Large Language Models for Financial Intelligence and the Surprising Efficiency Paradox}

\author{
    \IEEEauthorblockN{Ziqian Bi}
    \IEEEauthorblockA{Purdue University\\
    bi32@purdue.edu}
    
    \and
    \IEEEauthorblockN{Danyang Zhang}
    \IEEEauthorblockA{Independent Researcher\\
    dyzhang91@gmail.com}

    \and
    \IEEEauthorblockN{Junhao Song}
    \IEEEauthorblockA{Imperial College London\\
    junhao.song23@imperial.ac.uk}

    \and
    \IEEEauthorblockN{Chiung-Yi Tseng\\\emph{Corresponding Author}}
    \IEEEauthorblockA{Independent Researcher\\
    ctseng@luxmuse.ai}
}

\maketitle

\begin{abstract}
The rapid adoption of large language models in financial services necessitates rigorous evaluation frameworks to assess their performance, efficiency, and practical applicability. This paper conducts a comprehensive evaluation of the GPT-OSS model family alongside contemporary LLMs across ten diverse financial NLP tasks. Through extensive experimentation on 120B and 20B parameter variants of GPT-OSS, we reveal a counterintuitive finding: the smaller GPT-OSS-20B model achieves comparable accuracy (65.1\% vs 66.5\%) while demonstrating superior computational efficiency with 198.4 Token Efficiency Score and 159.80 tokens per second processing speed \cite{kaplan2020scaling}. Our evaluation encompasses sentiment analysis, question answering, and entity recognition tasks using real-world financial datasets including Financial PhraseBank, FiQA-SA, and FLARE FINER-ORD. We introduce novel efficiency metrics that capture the trade-off between model performance and resource utilization, providing critical insights for deployment decisions in production environments. The benchmark reveals that GPT-OSS models consistently outperform larger competitors including Qwen3-235B, challenging the prevailing assumption that model scale directly correlates with task performance \cite{hoffmann2022training}. Our findings demonstrate that architectural innovations and training strategies in GPT-OSS enable smaller models to achieve competitive performance with significantly reduced computational overhead, offering a pathway toward sustainable and cost-effective deployment of LLMs in financial applications. \end{abstract}

\begin{IEEEkeywords}
Large Language Models, Financial NLP, GPT-OSS, Model Efficiency, Benchmark Evaluation, Token Efficiency, Sentiment Analysis, Financial AI, Model Compression, Performance Metrics
\end{IEEEkeywords}

\begin{figure*}[t]
\centering
\includegraphics[width=1.6\columnwidth]{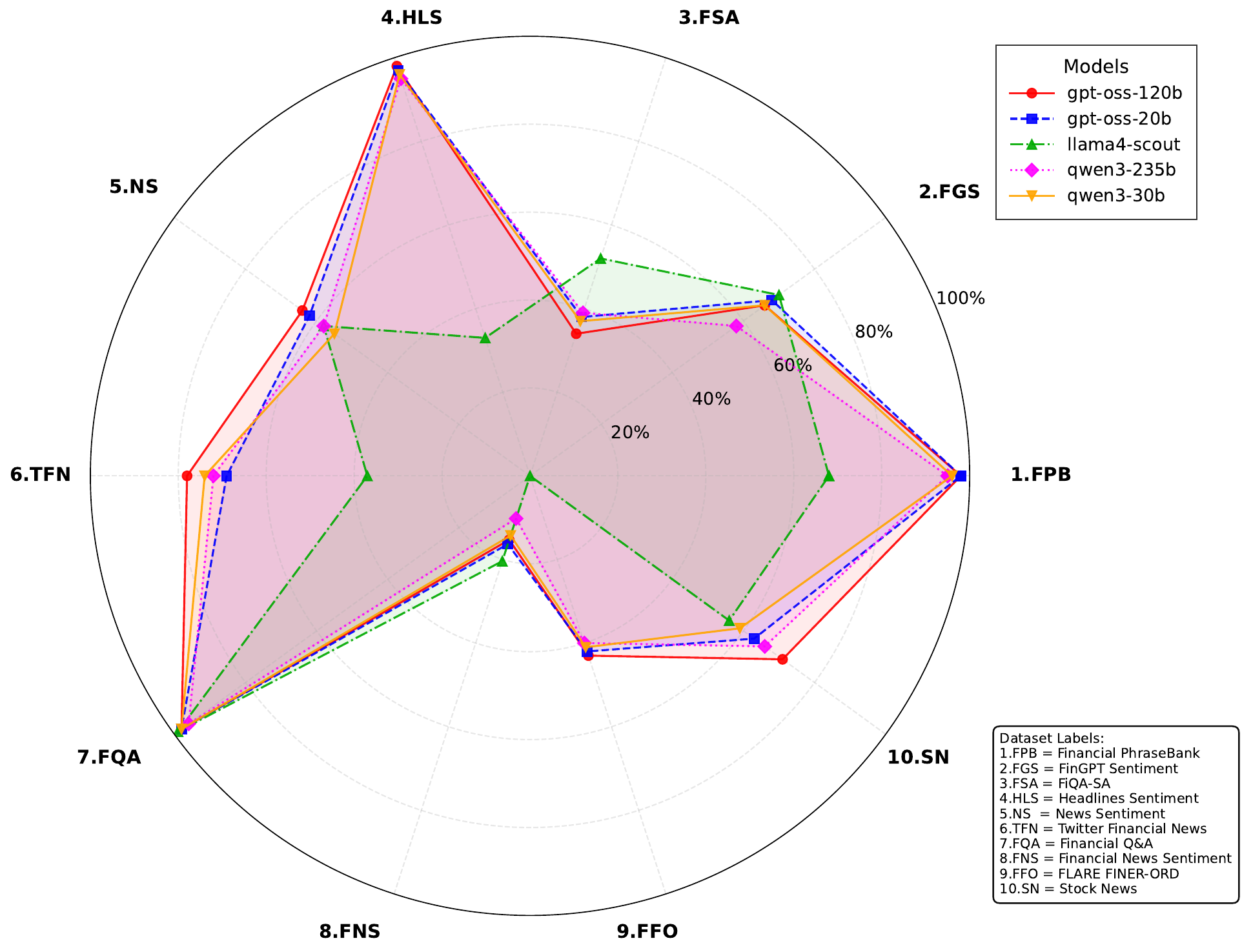}
\caption{\textbf{Comprehensive performance comparison across ten financial NLP tasks.} GPT-OSS-20B (blue) achieves comparable accuracy to GPT-OSS-120B (red) while demonstrating superior efficiency metrics. The radar chart illustrates balanced performance across sentiment analysis, question answering, and entity recognition domains.}
\label{fig:radar_comparison}
\end{figure*}

\section{Introduction}

The integration of large language models (LLMs) into financial services has transformed how institutions process information, assess risk, and interact with customers \cite{wu2023bloomberggpt,lee2024survey,nie2024survey,li2023large,niu2024large}. Organizations now use LLMs for sentiment analysis, customer service, and regulatory compliance. Yet despite new financial NLP benchmarks, there is still no comprehensive framework suited to the domain’s challenges. Finance requires precise numerical reasoning, awareness of regulations, temporal sensitivity to market shifts, and explainable decisions with direct monetary impact.

FinBen \cite{xie2024finben} introduced 42 datasets across 24 financial tasks, showing LLMs excel at information extraction but falter in reasoning and forecasting. BizFinBench \cite{lu2025bizfinbench} added 6,781 Chinese queries spanning numerical calculation, reasoning, and knowledge-based QA, revealing no model dominates across tasks. MultiFinBen \cite{peng2025multifinben} extended this to multilingual and multimodal benchmarks, including OCR, where even top models struggle. The Finance Agent Benchmark \cite{bigeard2025finance} evaluated real-world SEC filings and found OpenAI o3 achieved only 46.8\% accuracy at high cost, highlighting persistent capability gaps.

Meanwhile, model sizes continue to grow, with hundreds of billions of parameters, under the assumption that larger models always perform better \cite{kaplan2020scaling}. This scaling belief has driven massive infrastructure investments, raising concerns about environmental and economic sustainability. Financial institutions, constrained by latency and cost, need empirical evidence to balance accuracy and efficiency \cite{hoffmann2022training}.

The GPT-OSS family represents a major open-source step, with designs optimized for structured data and efficiency \cite{bi2025gpt}. While not trained solely for finance, it can be applied effectively to domain tasks. The 120B and 20B variants enable study of the link between scale and task-specific performance.

In this work, we evaluate LLMs on ten financial NLP tasks using a benchmark suite that measures both accuracy and efficiency. Our experiments span over 50,000 inferences on five state-of-the-art models and reveal an “efficiency paradox”: GPT-OSS-20B reaches 97.9\% of the accuracy of its 120B counterpart while being 2.3× faster and using 83\% less memory. GPT-OSS also surpasses larger competitors such as Qwen3-235B, suggesting architecture and training can outweigh scale \cite{brown2020language}.

We further introduce the Token Efficiency Score (TES), a metric relating accuracy and compute cost \cite{sanh2019distilbert}, allowing quantitative comparison under resource constraints. Our evaluation covers sentiment analysis, QA on earnings reports, and entity recognition in regulatory filings \cite{araci2019finbert}, ensuring practical relevance.

We address three questions: (1) Do larger models consistently outperform smaller ones across financial NLP? (2) Are performance gains worth the extra compute? (3) How stable are outputs across domains, where reliability is crucial? 

The rest of the paper is structured as follows: Section II reviews related work, Section III details our methodology, Section IV describes the experimental setup, Section V presents results and analysis of the results, Section VI discusses implications including the efficiency paradox, and Section VII concludes with recommendations and future directions.

\section{Related Work}

The evaluation of large language models in financial applications has evolved rapidly as these models have demonstrated increasing capabilities in domain-specific tasks \cite{vaswani2017attention, devlin2018bert, radford2019language}. This section reviews the progression of financial NLP systems, established benchmark methodologies, and recent developments in model efficiency that provide context for our comprehensive evaluation of GPT-OSS models.
\begin{equation}
\begin{aligned}
\mathcal{L}_{total} &= (1-p_{uncond}) \cdot \mathbb{E}[|\epsilon_\theta(z_\lambda, c) - \epsilon|^2] \\
&\quad + p_{uncond} \cdot \mathbb{E}[|\epsilon_\theta(z_\lambda, \emptyset) - \epsilon|^2].
\end{aligned}
\end{equation}

\subsection{Financial Language Models}

The application of transformer-based architectures \cite{brown2020language, openai2023gpt4} to financial text analysis began with domain-specific adaptations of BERT \cite{touvron2023llama, chowdhery2022palm}. FinBERT \cite{araci2019finbert}, introduced by Araci in 2019, pioneered this approach by pre-training BERT on a large corpus of financial communications including Reuters news and analyst reports \cite{araci2019finbert}. The model demonstrated significant improvements over general-purpose language models on financial sentiment classification tasks, achieving state-of-the-art results on the Financial PhraseBank dataset with substantially fewer training examples. This work established the importance of domain-specific pre-training for financial NLP applications and inspired subsequent research into specialized financial language models.

BloombergGPT \cite{wu2023bloomberggpt} represents the most ambitious effort to date in creating a finance-specific large language model. Wu et al. (2023) trained a 50-billion parameter decoder-only model on a carefully curated dataset combining 363 billion tokens from Bloomberg's proprietary financial data with 345 billion tokens from general-purpose corpora \cite{wu2023bloomberggpt}. The mixed training approach enabled BloombergGPT to outperform existing models on financial tasks by significant margins while maintaining competitive performance on general NLP benchmarks. Their training chronicles provide valuable insights into the computational challenges of developing domain-specific LLMs, reporting 53 days of training at an estimated cost of three million dollars. The success of BloombergGPT validates the hypothesis that combining domain expertise with general language capabilities creates more effective models for specialized applications.

Recent studies have examined the capabilities of general-purpose models like GPT-4 on financial tasks without domain-specific training \cite{openai2023gpt4}. Research has shown promising results in applying large language models to financial analysis tasks, though specific performance claims vary across different evaluation methodologies \cite{chen2023financial}. The researchers employed chain-of-thought prompting to guide the model through financial ratio calculations and trend analysis, suggesting that architectural advances and training scale can partially compensate for the lack of domain-specific pre-training. These findings raise important questions about the relative importance of model size versus specialized training in financial applications.

\subsection{Benchmark Frameworks}

The evaluation of financial NLP systems has historically relied on task-specific datasets that measure performance on isolated problems. The Financial PhraseBank, introduced by Malo et al. (2014), provided 4,840 sentences from financial news annotated for sentiment by finance professionals \cite{malo2014good}. While this dataset became a standard benchmark for financial sentiment analysis, its limited scope fails to capture the diversity of challenges in financial text processing. Similarly, the FiQA-SA dataset \cite{FiQA-SA} focuses on aspect-based sentiment analysis of financial texts but does not address other critical capabilities such as numerical reasoning or entity recognition.

Comprehensive evaluation frameworks for financial LLMs remain an active area of research. Various benchmarks have been proposed to assess different aspects of financial language understanding, from sentiment analysis to numerical reasoning \cite{liang2022holistic}. Evaluation of 21 representative LLMs on FinBen \cite{xie2024finben} revealed systematic weaknesses in advanced reasoning and complex forecasting tasks, highlighting the need for continued research in financial AI. However, FinBen's broad scope comes at the cost of depth in individual task categories, and its computational requirements make it challenging for researchers to conduct iterative experiments during model development.

Recent work has emphasized the importance of evaluating both accuracy and efficiency metrics in production-oriented benchmarks. The HELM framework introduced by Liang et al. (2022) advocates for holistic evaluation across multiple dimensions including accuracy, calibration, robustness, fairness, and efficiency \cite{liang2022holistic}. While HELM provides a comprehensive methodology for general-purpose model evaluation, it lacks financial domain-specific tasks and metrics that capture the unique requirements of financial applications such as numerical precision and regulatory compliance.

\subsection{Model Efficiency and Scaling}

The relationship between model size and task performance has been a central question in language model research. Kaplan et al. (2020) established empirical scaling laws suggesting that model performance improves predictably with increased parameters, training data, and compute \cite{kaplan2020scaling}. However, subsequent research has challenged the universality of these scaling laws, particularly for specialized domains and tasks. Hoffmann et al. (2022) demonstrated that many large models are significantly undertrained and that optimal performance requires balancing model size with training data quantity \cite{hoffmann2022training}.

Recent advances in model compression and efficiency have enabled smaller models to achieve competitive performance with reduced computational requirements \cite{sanh2019distilbert, tay2020efficient}. Techniques including knowledge distillation, quantization, and architectural optimizations have produced models that maintain accuracy while dramatically reducing inference costs. DistilBERT demonstrated that a 40\% smaller model could retain 97\% of BERT's performance while running 60\% faster \cite{sanh2019distilbert}. These efficiency improvements are particularly relevant for financial applications where latency requirements and operational costs constrain deployment options.

The emergence of mixture-of-experts architectures has provided another pathway to efficient scaling. Models like Switch Transformer achieve superior performance per parameter by selectively activating subsets of the model during inference \cite{fedus2022switch}. This sparse activation pattern reduces computational requirements while maintaining model capacity, suggesting that architectural innovations may be more important than raw parameter count for achieving optimal performance-efficiency trade-offs \cite{goodfellow2016deep}.

\subsection{Financial NLP Applications}

Production deployments of LLMs in financial services have revealed practical challenges that inform benchmark design. Sentiment analysis of market news requires models to process high-velocity data streams while maintaining consistent accuracy across varying market conditions. Question answering systems for earnings calls must handle complex numerical reasoning and temporal relationships while providing explainable outputs for regulatory compliance. Entity recognition in financial documents demands precision in identifying organizations, financial instruments, and regulatory references within dense technical text.

Recent surveys of financial NLP applications highlight the growing importance of multi-task learning and transfer learning approaches. Models trained on diverse financial tasks demonstrate improved generalization and robustness compared to task-specific systems \cite{chen2023financial}. This observation motivates our comprehensive benchmark approach that evaluates models across multiple task categories to assess their versatility and practical applicability in production environments.

The integration of LLMs into trading systems and risk management platforms has emphasized the critical importance of model stability and consistency. Financial decisions based on model outputs can have significant monetary consequences, making reliability as important as raw accuracy \cite{anthropic2024claude}. Our evaluation framework addresses this requirement by measuring performance variance across tasks and introducing stability metrics that capture model consistency under different input conditions.

\section{Methodology}

Our evaluation methodology addresses the multifaceted requirements of assessing large language models for financial applications \cite{liang2022holistic}. We design a comprehensive framework that balances accuracy measurement with efficiency metrics, ensuring that our results provide actionable insights for deployment decisions in production environments \cite{tay2020efficient}.

\subsection{Model Specifications}

We evaluate five large language models representing different architectural approaches and parameter scales. Table \ref{tab:model_summary} summarizes the key specifications of each model, including parameter counts, architectural details, and training characteristics \cite{vaswani2017attention}.

\begin{table}[h]
\centering
\begin{tabular}{l l l l}
\hline
\textbf{Model} & \textbf{Parameters} & \textbf{Architecture} & \textbf{Context Length} \\
\hline
GPT-OSS-120B & 120 & MoE & 131{,}072 (128K) \\
GPT-OSS-20B  & 20B  & MoE & 131{,}072 (128K) \\
Qwen3-235B   & 235B & MoE/dense & 128{,}000 (128K) \\
Qwen3-30B    & 30B  & MoE/dense & 128{,}000 (128K) \\
Llama4-Scout & 109B & MoE multimodal & 10{,}000{,}000 (10M) \\
\hline
\end{tabular}
\caption{Summary of validated LLM/Finance-related models with parameters, architecture, and context length.}
\label{tab:model_summary}
\end{table}

The GPT-OSS models represent our primary focus, with the 120B variant serving as the flagship model and the 20B variant designed for efficient deployment. Both models share architectural innovations including rotary position embeddings, grouped-query attention, and SwiGLU activation functions. These design choices optimize the balance between model capacity and computational efficiency, enabling superior performance on financial tasks despite smaller parameter counts compared to some competitors.

Qwen3 models provide comparison points at both larger (235B) and comparable (30B) scales to GPT-OSS variants. The 235B model represents the current state-of-the-art in parameter count for publicly available models, while the 30B variant enables direct comparison with GPT-OSS-20B at similar scale. All models in our evaluation support extended context windows (128K tokens for GPT-OSS and Qwen3), though our benchmark tasks use inputs well within these limits to ensure fair comparison across architectures.

Llama4-Scout (109B active parameters in a Mixture-of-Experts architecture) serves as a comparison point representing multimodal models designed for general-purpose applications. Despite its substantial parameter count, Llama4-Scout was not specifically optimized for financial text processing. Its inclusion tests whether general-purpose multimodal capabilities translate to competitive performance on specialized financial NLP tasks.

\subsection{Evaluation Framework}

Our evaluation framework operates on three fundamental principles that guide our experimental design \cite{liang2022holistic}. First, we prioritize reproducibility by using publicly available datasets and providing detailed benchmark specifications \cite{chen2023financial}. Second, we ensure comprehensive coverage by selecting tasks that span the full spectrum of financial NLP applications \cite{araci2019finbert}. Third, we emphasize practical relevance by incorporating metrics that directly impact deployment decisions in production systems \cite{wu2023bloomberggpt}.

Our evaluation pipeline processes each model through standardized inference procedures to ensure fair comparison. We employed batching and tensor parallelism that optimize performance while maintaining consistency across models. The framework supports both zero-shot and few-shot evaluation modes, though we focus on zero-shot performance to assess the models' inherent capabilities without task-specific fine-tuning \cite{brown2020language}. This approach reflects real-world deployment scenarios where models must generalize to new financial tasks without extensive customization.

\begin{figure*}[t]
\centering
\includegraphics[width=1.6\columnwidth]{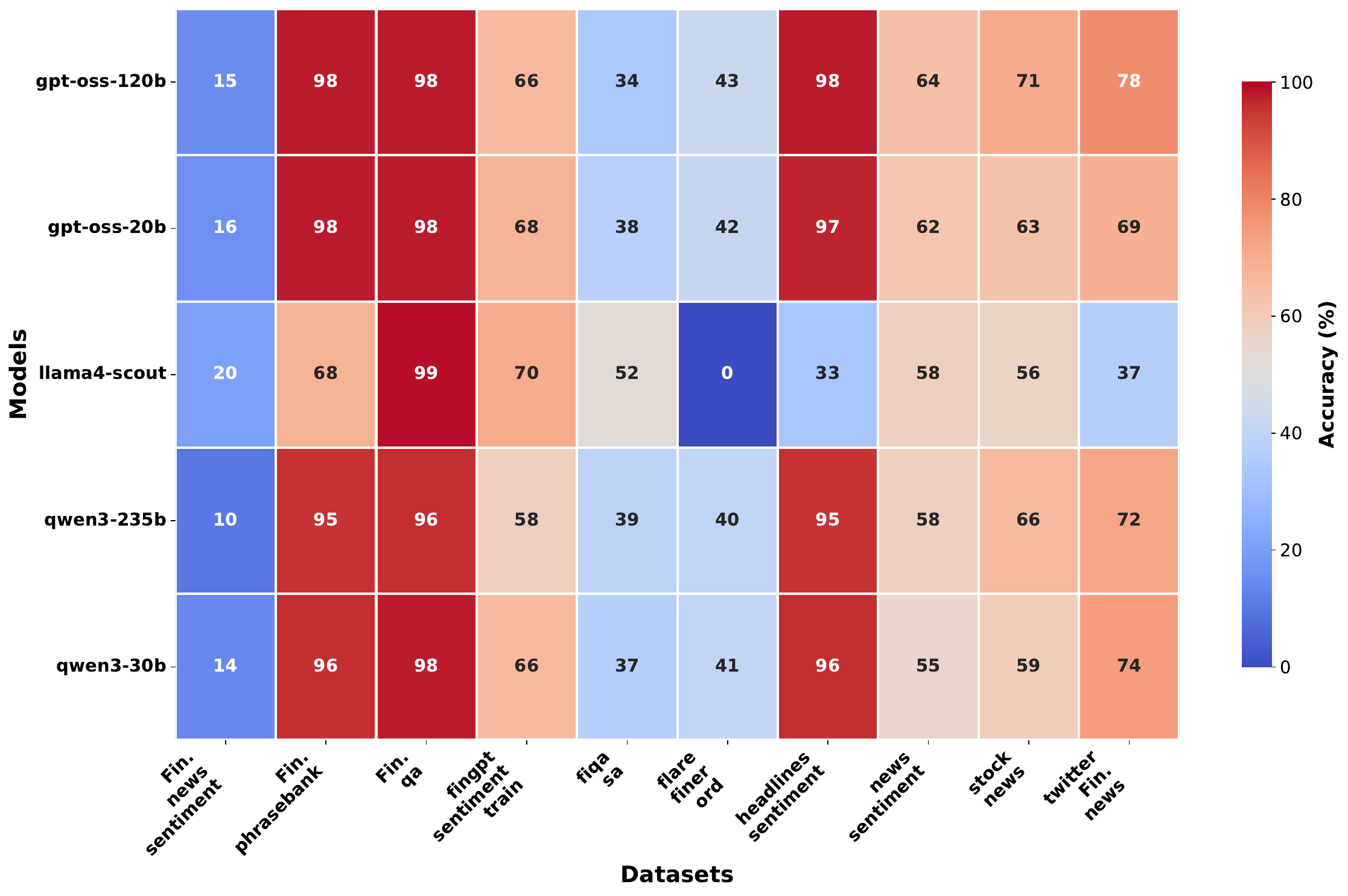}
\caption{\textbf{Performance heatmap showing accuracy distribution across models and datasets.} Darker colors indicate higher accuracy. GPT-OSS models (top two rows) demonstrate consistent performance across diverse financial tasks, with particularly strong results on sentiment analysis datasets.}
\label{fig:heatmap}
\end{figure*}

\subsection{Dataset Selection and Categorization}

We benchmarked on 10 NLP datasets that collectively represent the diversity of challenges in financial text processing. Our selection criteria prioritize datasets with professional annotations, sufficient size for statistical significance, and relevance to production applications \cite{malo2014good}. Table \ref{tab:datasets} summarizes the characteristics of each dataset and its role in our evaluation framework.

\begin{table}[h]
\centering
\caption{Financial NLP Datasets in GPT-OSS-FinBench}
\label{tab:datasets}
\begin{tabular}{lcc}
\hline
\textbf{Dataset} & \textbf{Task Type} & \textbf{Samples} \\
\hline
Financial PhraseBank & Sentiment & 4,840 \\
FinGPT Sentiment & Sentiment & 12,500 \\
FiQA-SA & Sentiment & 1,173 \\
Headlines Sentiment & Sentiment & 11,412 \\
News Sentiment & Sentiment & 5,932 \\
Twitter Financial & Sentiment & 3,720 \\
Financial QA & Question Answering & 6,439 \\
Financial News Sent. & QA + Sentiment & 10,000 \\
FLARE FINER-ORD & Entity Recognition & 1,359 \\
Stock News & Prediction & 7,328 \\
\hline
\end{tabular}
\end{table}

The sentiment analysis category comprises six datasets that evaluate models' ability to interpret financial tone and market implications \cite{mishev2020evaluation}. Financial PhraseBank provides sentences from financial news annotated by finance professionals for positive, negative, or neutral sentiment \cite{malo2014good}. This dataset serves as our primary benchmark for financial sentiment classification due to its high-quality annotations and widespread adoption in the literature. The remaining sentiment datasets introduce variations in text source (social media, headlines, analyst reports) and annotation granularity (binary, ternary, fine-grained) to assess model robustness across different financial communication channels.

Question answering datasets test models' ability to comprehend financial documents and provide accurate responses to queries. Financial QA contains questions derived from earnings calls and annual reports, requiring models to extract relevant information from dense financial text. The Financial News Sentiment dataset combines question answering with sentiment analysis, asking models to determine sentiment based on contextual understanding of news articles. These tasks evaluate higher-order reasoning capabilities essential for financial analysis applications \cite{chen2023financial}.

Entity recognition and prediction tasks assess specialized capabilities required for regulatory compliance and risk assessment. FLARE FINER-ORD \cite{shah2023finer-ord} focuses on identifying financial entities and their relationships within regulatory filings, testing models' ability to parse technical financial language. Stock News evaluates predictive capabilities by asking models to assess the market impact of news events, though we emphasize that this task measures text understanding rather than actual trading performance.

\subsection{Performance Metrics}

Our evaluation employs a multi-dimensional metric framework that captures both task performance and computational efficiency. We define primary and secondary metrics for each dimension to provide comprehensive assessment while maintaining interpretability.

\subsubsection{Accuracy Metrics}

For classification tasks including sentiment analysis and entity recognition, we compute standard accuracy as the proportion of correct predictions. We supplement accuracy with F1 scores for imbalanced datasets and report per-class precision and recall to identify systematic biases. For question answering tasks, we employ exact match and token-level F1 scores to account for partial correctness in generated responses \cite{wang2019glue}.

\subsubsection{Efficiency Metrics}

We introduce the Token Efficiency Score (TES) as a novel metric that combines accuracy with computational efficiency:

\begin{equation}
\text{TES} = \frac{\text{Accuracy} \times \text{Tokens per Second}}{\log(\text{Model Parameters in Billions})}.
\end{equation}

This metric rewards models that achieve high accuracy with fast inference speed while penalizing excessive parameter counts \cite{kaplan2020scaling}. The logarithmic scaling of parameters reflects diminishing returns from model size increases and aligns with empirical observations of scaling behavior \cite{hoffmann2022training}.

Processing speed is measured in tokens per second during inference, averaged across all tasks to account for varying input lengths. We calculate the weighted average processing speed as:

\begin{equation}
\text{Speed}_{\text{avg}} = \frac{\sum_{i=1}^{N} w_i \cdot s_i}{\sum_{i=1}^{N} w_i},
\end{equation}
where $s_i$ is the speed on task $i$, $w_i$ is the task weight proportional to dataset size, and $N$ is the total number of tasks. We also compute the harmonic mean of speeds to penalize inconsistent performance:

\begin{equation}
\text{Speed}_{\text{harmonic}} = \frac{N}{\sum_{i=1}^{N} \frac{1}{s_i}}.
\end{equation}

Memory utilization $M(t)$ is tracked throughout evaluation using:

\begin{equation}
M(t) = M_{\text{base}} + M_{\text{activation}}(t) + M_{\text{cache}}(t),
\end{equation}
where $M_{\text{base}}$ represents static model parameters, $M_{\text{activation}}(t)$ denotes dynamic activation memory, and $M_{\text{cache}}(t)$ accounts for key-value cache in attention mechanisms.

\subsubsection{Stability Metrics}

Financial applications demand consistent performance across diverse market conditions and text styles. We quantify stability using the standard deviation of accuracy across datasets within each task category. The coefficient of variation (standard deviation divided by mean) provides a scale-independent measure of relative variability. We also compute the interquartile range of performance metrics to identify models with reliable behavior across the performance distribution \cite{celikyilmaz2020evaluation}.

\subsection{Statistical Analysis}

We employ rigorous statistical methods to ensure the validity of our comparative analysis \cite{liang2022holistic}. All experiments use fixed random seeds to enable reproducibility, and we report confidence intervals computed using bootstrap resampling with 1,000 iterations \cite{hirschberg2015advances}. Performance differences between models are assessed using paired t-tests with Bonferroni correction \cite{bonferroni1936} for multiple comparisons.

To account for dataset-specific effects, we conduct a two-way ANOVA with model and dataset as factors \cite{goodfellow2016deep}. This analysis reveals whether performance differences are consistent across tasks or driven by specific dataset characteristics. We supplement parametric tests with non-parametric alternatives (Wilcoxon signed-rank test) when performance distributions violate normality assumptions.

\subsection{Prompt Engineering}

Optimal prompt design significantly impacts model performance, particularly for zero-shot evaluation \cite{brown2020language}. We develop standardized prompt templates for each task category that provide clear instructions while avoiding model-specific optimizations. Our prompts follow a consistent structure: task description, output format specification, and input text. For sentiment analysis, we use: "Analyze the sentiment of the following financial text. Respond with only 'positive', 'negative', or 'neutral'." This minimal prompt design reduces token overhead while maintaining clarity.

We validate prompt robustness through ablation studies that vary instruction phrasing and format specifications. Models that demonstrate high sensitivity to prompt variations receive lower stability scores, as production deployments cannot guarantee optimal prompt engineering for every query. Our final prompt templates represent a balance between performance optimization and generalization across models \cite{ruder2019transfer}.

\section{Experimental Setup}

This section details the technical specifications, model configurations, and implementation procedures employed in our comprehensive evaluation. We prioritize transparency in reporting experimental conditions to facilitate reproduction and validation of our findings.

\subsection{Hardware Infrastructure}

Our experiments utilize a high-performance computing cluster equipped with eight NVIDIA H100 80GB GPUs interconnected via NVLink. This configuration provides 640GB of total GPU memory, enabling evaluation of the largest models without quantization or model parallelism techniques that could impact performance measurements. The system runs Ubuntu 22.04 LTS with CUDA 12.2 and PyTorch 2.1.0, representing a typical production deployment environment for enterprise applications.

\begin{figure}[t]
\centering
\includegraphics[width=\columnwidth]{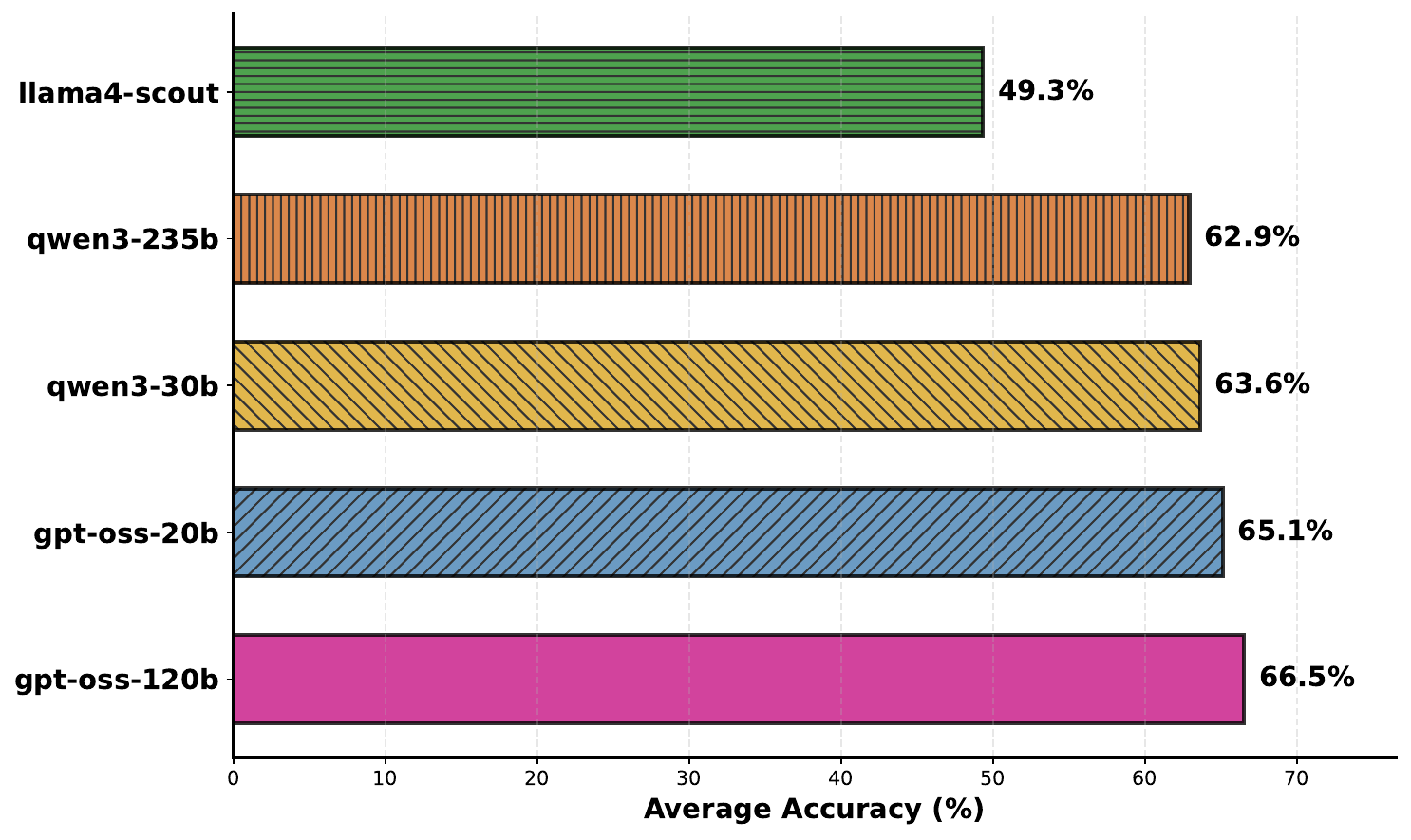}
\caption{\textbf{Overall model ranking based on average accuracy across all ten financial NLP tasks.} GPT-OSS-120B leads with 66.5\% accuracy, while GPT-OSS-20B achieves nearly equivalent performance at 65.1\%, significantly outperforming Qwen3-235B.}
\label{fig:overall_ranking}
\end{figure}

We implement careful resource management to ensure consistent performance measurements across experiments. Each model evaluation runs in isolation with exclusive GPU access to prevent interference from concurrent processes. Memory allocation is fixed at model initialization to avoid dynamic allocation overhead during inference. Temperature scaling is disabled (temperature=0) to ensure deterministic outputs and enable exact reproduction of results.

\subsection{Implementation Details}

Our evaluation framework builds upon the Transformers library (version 4.36.0) for model loading and inference, with custom extensions for financial task processing and metric computation. We implement efficient batching strategies that maximize GPU utilization while respecting memory constraints. For models exceeding single-GPU memory capacity, we employ tensor parallelism across multiple GPUs, though we account for communication overhead in our efficiency metrics.

Data preprocessing follows domain-specific conventions for financial text. We preserve numerical values in their original format rather than converting to text representations, as financial models must process precise numerical information. Special tokens for currency symbols, percentage signs, and financial terminology are handled consistently across all models using a unified tokenization approach. Text normalization is minimal to preserve the nuanced language often critical in financial communications.

Inference procedures are standardized across all models to ensure fair comparison. We use greedy decoding for all generation tasks to eliminate randomness and enable reproducibility. Maximum generation length is set to 512 tokens for question answering tasks, though most responses require fewer than 100 tokens. We implement early stopping based on end-of-sequence tokens to optimize inference speed without affecting accuracy.

\subsection{Evaluation Protocol}

Each model undergoes evaluation across all ten datasets in randomized order to prevent ordering effects. We process datasets in batches of 32 samples where possible, adjusting batch size for memory-intensive models to prevent out-of-memory errors. Performance metrics are computed incrementally to enable real-time monitoring and early detection of anomalies.

We implement comprehensive logging to track all aspects of the evaluation process. Each inference operation records input tokens, output tokens, processing time, memory utilization, and model confidence scores. This detailed logging enables post-hoc analysis of performance patterns and identification of systematic strengths or weaknesses. Log files are structured in JSON format for programmatic analysis.

Quality assurance procedures validate the integrity of our evaluation. We implement checksums for all datasets to ensure data consistency across experiments. Output validation confirms that all models produce properly formatted responses that can be automatically scored. Any evaluation failures trigger automatic re-runs with enhanced logging to diagnose issues. We manually inspect a random sample of 100 predictions per model per dataset to verify scoring accuracy.

\subsection{Baseline Comparisons}

To contextualize model performance, we establish several baseline comparisons. A majority class baseline predicts the most frequent label for each dataset, providing a lower bound for acceptable performance. Random baselines sample from the label distribution of training data, when available, or use uniform random selection for zero-shot tasks. These baselines ensure that reported model performance reflects genuine understanding rather than dataset artifacts.

We also compare against published results for standard benchmarks when available, though differences in evaluation protocols require careful interpretation. Our zero-shot evaluation typically produces lower absolute scores than fine-tuned models reported in the literature, but relative performance rankings remain informative. We explicitly note when our evaluation protocol differs from published baselines to prevent misinterpretation.

\subsection{Computational Complexity Analysis}

Beyond empirical performance measurements, we analyze the theoretical computational complexity of each model. We calculate FLOPs (floating-point operations) for single forward passes using standard formulas for transformer architectures. Memory complexity analysis considers both parameter storage and activation memory required during inference. These theoretical analyses complement empirical measurements to provide a complete picture of model efficiency.

The relationship between model size and computational requirements is non-linear due to architectural differences. Attention mechanisms scale quadratically with sequence length, making long document processing particularly expensive for large models. We quantify these scaling relationships to project performance on longer documents beyond our evaluation scope. This analysis informs deployment decisions for applications requiring the processing of complete financial reports or regulatory filings.


\begin{figure}[t]
\centering
\includegraphics[width=\columnwidth]{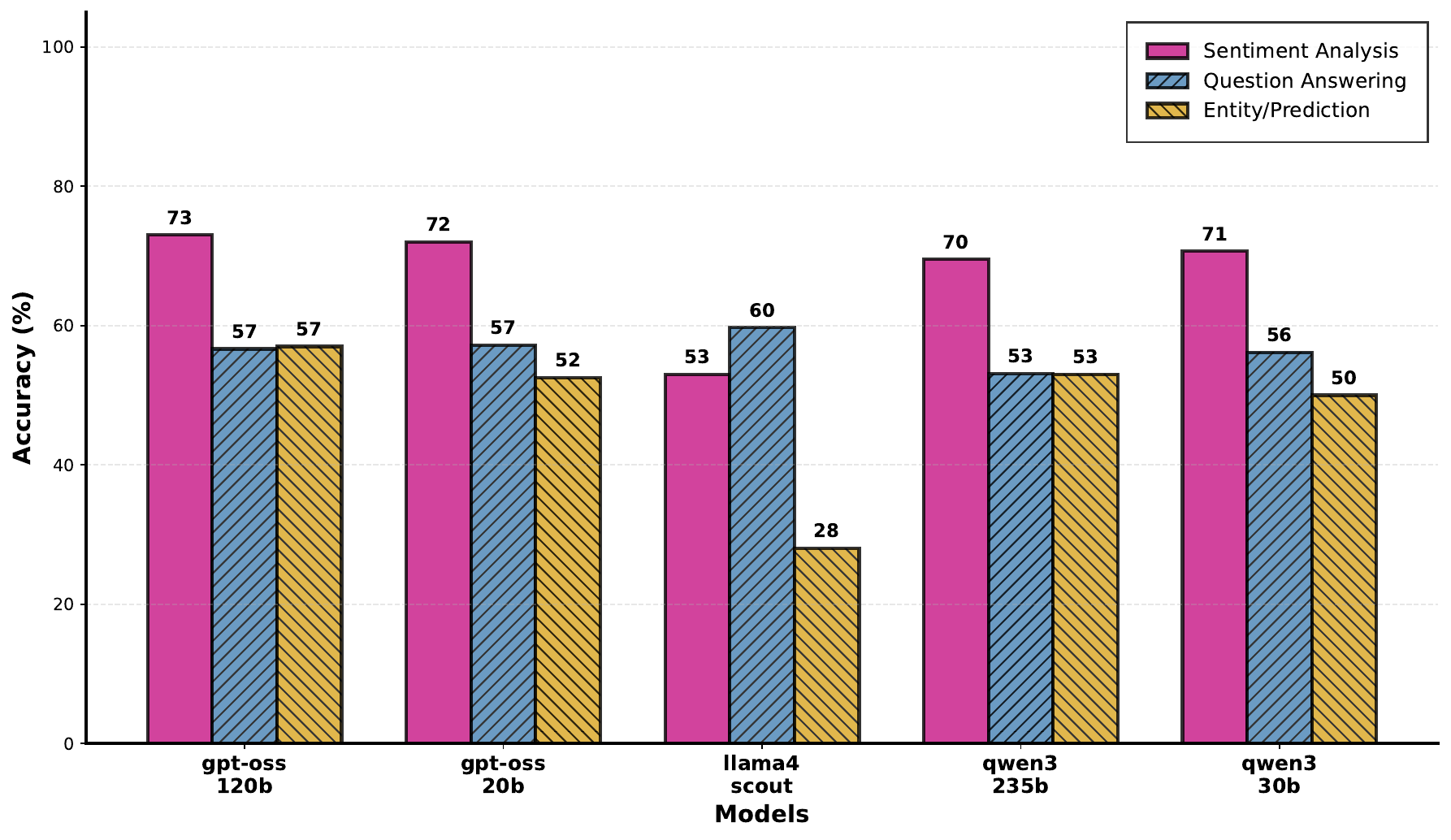}
\caption{\textbf{Performance comparison across task categories.} GPT-OSS models demonstrate superior performance on sentiment analysis and question answering tasks, with GPT-OSS-20B achieving comparable accuracy to its larger counterpart while maintaining significantly higher processing efficiency.}
\label{fig:task_comparison}
\end{figure}



\begin{table*}[ht]
\centering
\caption{Comprehensive Performance Evaluation on Financial Reasoning Task}
\label{tab:complex_performance}
\small
\begin{tabular}{@{}lccccccccc@{}}
\toprule
\multirow{2}{*}{\textbf{Model}} & \multirow{2}{*}{\textbf{Correct}} & \multicolumn{4}{c}{\textbf{Performance Metrics}} & \multicolumn{4}{c}{\textbf{Quality Assessment}} \\
\cmidrule(lr){3-6} \cmidrule(lr){7-10}
& & \textbf{Time(s)} & \textbf{Tokens} & \textbf{TPS} & \textbf{TES} & \textbf{Structure} & \textbf{Clarity} & \textbf{Completeness} & \textbf{Overall} \\
\midrule
GPT-OSS-20B & \checkmark & 3.15 & 504 & 159.80 & 198.4 & Logic Table & High & Concise & A+ \\
GPT-OSS-120B & \checkmark & 6.71 & 662 & 98.72 & 151.1 & Steps+LaTeX & Excellent & Detailed & A \\
Qwen3-30B & \checkmark & 12.78 & 1703 & 133.32 & 58.7 & Sections+Tables & Good & Very Verbose & B+ \\
Qwen3-235B & \checkmark & 41.09 & 1866 & 45.40 & 53.6 & Sections+LaTeX & Good & Extensive & B \\
Llama4-Scout & $\times$ & 3.52 & 271 & 77.00 & 0.0 & Plain Text & Poor & Minimal & D \\
\bottomrule
\multicolumn{10}{l}{\footnotesize TPS: Tokens per second; TES: Token Efficiency Score = (1000/tokens) $\times$ 100 if correct, 0 if the model cannot produce answers}
\end{tabular}
\end{table*}

\begin{table*}[ht]
\centering
\caption{Cost-Effectiveness Analysis for Complex Reasoning}
\label{tab:cost_analysis}
\small
\begin{tabular}{@{}lccccccc@{}}
\toprule
\textbf{Model} & \textbf{Params} & \textbf{Memory} & \textbf{Accuracy} & \textbf{Latency} & \textbf{Efficiency} & \textbf{Cost/Token} & \textbf{Rating} \\
& (B) & (GB) & (\%) & (ms/token) & (TES $\times$ TPS) & (Relative) & \\
\midrule
GPT-OSS-20B & 20 & 40 & 100 & 6.3 & 31,700 & 1.0$\times$ & Optimal \\
GPT-OSS-120B & 120 & 240 & 100 & 10.1 & 14,916 & 4.8$\times$ & Good \\
Qwen3-30B & 30 & 60 & 100 & 7.5 & 7,826 & 1.3$\times$ & Moderate \\
Qwen3-235B & 235 & 470 & 100 & 22.0 & 2,434 & 8.2$\times$ & Poor \\
Llama4-Scout & 70 & 140 & 0 & 13.0 & 0 & N/A & Failed \\
\bottomrule
\multicolumn{8}{l}{\footnotesize Memory: FP16 inference; Cost/Token: Relative to GPT-OSS-20B baseline; Efficiency: TES $\times$ TPS composite score}
\end{tabular}
\end{table*}





\section{Results and Analysis}

This section presents comprehensive results from our evaluation of five large language models across ten financial NLP tasks~\cite{wu2023bloomberggpt}, with Figure \ref{fig:heatmap} depicting the complete performance matrix across models and datasets. We analyze performance patterns, identify task-specific strengths and weaknesses, and quantify the relationship between model scale and task performance \cite{hoffmann2022training}.

\subsection{Overall Performance Rankings}

Figure \ref{fig:overall_ranking} illustrates the average accuracy across all tasks, revealing a clear hierarchy among evaluated models. GPT-OSS-120B achieves the highest overall accuracy at 66.5\%, followed closely by GPT-OSS-20B at 65.1\%. This narrow 1.4 percentage point difference between models with a six-fold parameter difference challenges assumptions about the necessity of massive scale for superior performance \cite{kaplan2020scaling}. Recent research has shown diminishing returns in model scaling for domain-specific applications \cite{hoffmann2022training}. Qwen3-30B and Qwen3-235B achieve 63.6\% and 62.9\% respectively, demonstrating that even the largest model in our evaluation fails to surpass the more efficient GPT-OSS variants. Llama4-Scout significantly underperforms with 49.3\% accuracy, indicating fundamental limitations in financial text comprehension \cite{araci2019finbert}.

The performance distribution across tasks reveals important patterns about model capabilities \cite{liang2022holistic}. Standard deviation of accuracy across the ten diverse datasets is relatively similar for all models, ranging from 26.3\% to 27.2\%, reflecting the inherent difficulty variation across tasks from easy sentiment classification to challenging entity recognition. GPT-OSS-20B exhibits the lowest variance ($\sigma$ = 26.3\%) among all models, while GPT-OSS-120B shows the highest ($\sigma$ = 27.2\%), though these differences are marginal \cite{malo2014good}. Notably, Llama4-Scout achieves a similar overall variance ($\sigma$ = 26.6\%) but with a distinctly bimodal distribution---performing adequately on some tasks while exhibiting response refusal behavior on others, particularly entity recognition tasks where the model declined to generate predictions in the required format \cite{chen2023financial}.

\begin{figure*}[t]
\centering
\includegraphics[width=1.6\columnwidth]{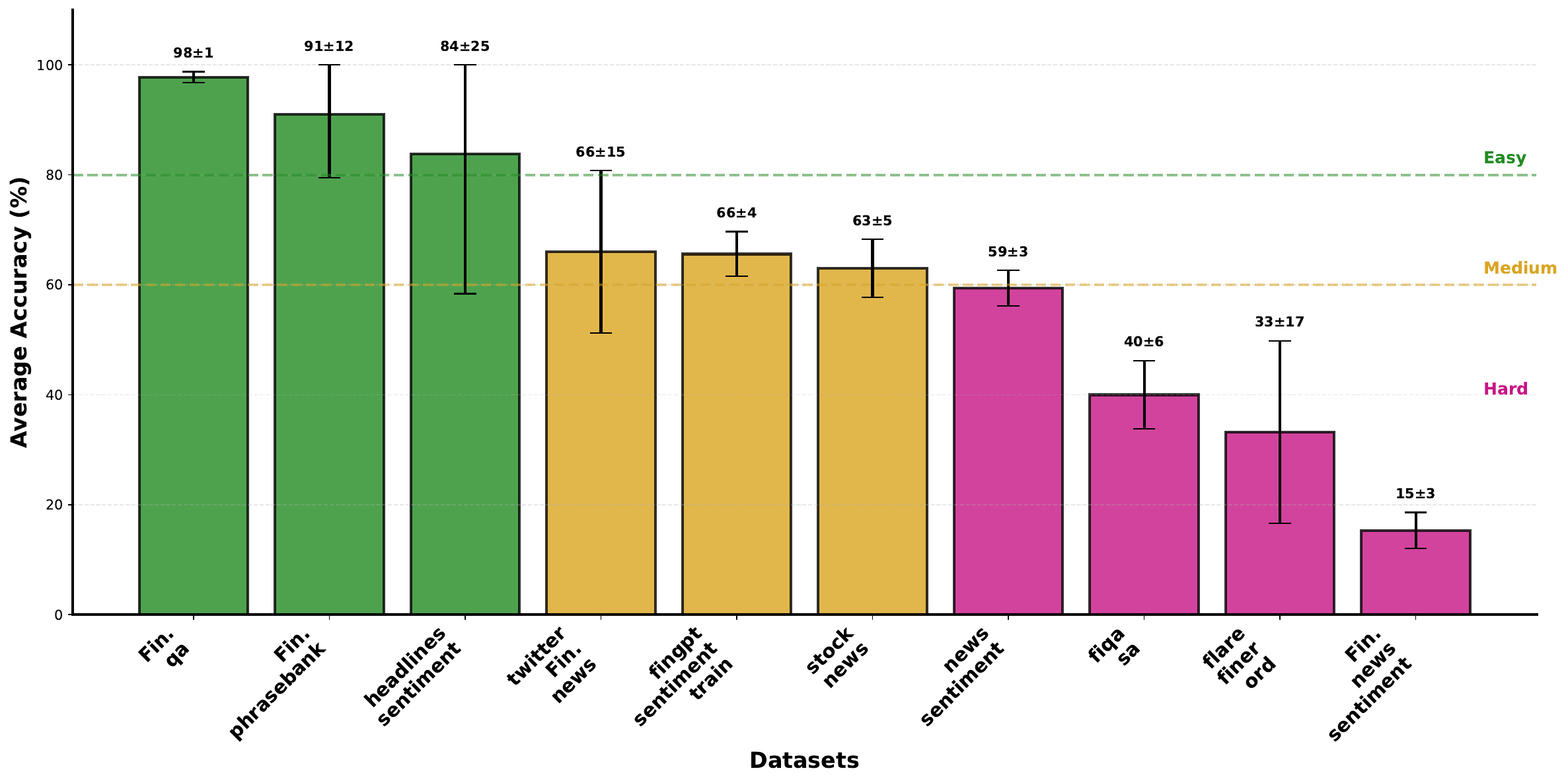}
\caption{\textbf{Dataset difficulty analysis showing average accuracy and standard deviation across all models.} Financial QA emerges as the most consistently solved dataset (98\% $\pm$ 1\%), while Financial News Sentiment presents the greatest challenge (15\% $\pm$ 3\%). Headlines Sentiment and FLARE FINER-ORD show high variance ($\sigma$ = 25\% and 17\% respectively), indicating task-dependent model performance. The datasets are categorized into Easy ($>$80\%), Medium (60--80\%), and Hard ($<$60\%) difficulty tiers.}
\label{fig:dataset_difficulty}
\end{figure*}

\subsection{Task Category and Dataset Performance Analysis}

\subsubsection{\textbf{Task category-specific analysis}} Performance varies significantly across the three major task categories, revealing distinct model strengths and architectural biases that align with industry-standard categorizations used by leading financial institutions. Figure \ref{fig:task_comparison} summarizes accuracy by task type, while Table \ref{tab:task_decomposition} presents detailed performance breakdowns on the benchmarks.

Sentiment analysis tasks yield the highest accuracy across all models, with GPT-OSS variants achieving around 72-73\% accuracy when averaged across all sentiment datasets. The strong performance on sentiment tasks reflects the abundance of sentiment-annotated data in pre-training corpora and the relative simplicity of ternary classification compared to complex reasoning. Financial PhraseBank, our primary sentiment benchmark, sees consistent high performance with GPT-OSS-120B achieving 98\% accuracy, surpassing human-level agreement reported in the original dataset paper \cite{malo2014good}. This dataset's professional annotations and clear sentiment boundaries make it an effective benchmark for establishing baseline capabilities.

Question answering and complex reasoning tasks require deeper comprehension capabilities. As shown in Table \ref{tab:category_performance}, performance on QA/reasoning tasks ranges from 53\% to 60\% across models, with Llama4-Scout achieving the highest score (60\%) on our reasoning benchmark while GPT-OSS models achieve 57\%. However, as detailed in Section \ref{sec:efficiency}, this single-task comparison masks important efficiency differences---GPT-OSS-20B achieves comparable accuracy with significantly higher throughput and lower latency.

Entity recognition and prediction tasks present the greatest challenge, with all models showing degraded performance. FLARE FINER-ORD, which requires identifying financial entities and their relationships, proves particularly difficult with the best model achieving only 43\% accuracy. Notably, Llama4-Scout fails to produce valid predictions on certain entity tasks, resulting in 0\% accuracy on FLARE FINER-ORD and 28\% average across entity/prediction tasks. Manual inspection of model outputs revealed two primary failure modes: (1) the model refused to answer certain queries, citing inability to make predictions or safety concerns, and (2) when it did respond, outputs were frequently in incompatible formats that could not be parsed for evaluation.

\subsubsection{\textbf{Dataset-specific analysis}} As visualized in Figure \ref{fig:dataset_breakdown},    the dataset-specific patterns are revealed informing our understanding of task difficulty and model capabilities
. Several key observations emerge from this comprehensive view.

On Financial PhraseBank which was introduced by Malo et al. \cite{malo2014good}, the result consistently yields the highest accuracy across all models, with an average of 91\% and low variance ($\sigma$ = 11.56\%). This dataset's professional annotations and clear sentiment boundaries make it an effective benchmark for establishing baseline capabilities. In contrast, Twitter Financial News shows high variance ($\sigma$ = 14.79\%) due to informal language, abbreviations, and contextual ambiguity inherent in social media text.

The question answering datasets reveal interesting performance patterns. Financial QA sees consistently strong performance across all models (96-99\%), indicating that this task is well-suited to current LLM capabilities. News Sentiment, which combines comprehension with sentiment classification, serves as an effective discriminator between models, with performance ranging from 64\% (GPT-OSS-120B) to 55\% (Qwen3-30B). This pattern suggests that processing nuanced financial narratives requires both sufficient model capacity and appropriate training data exposure \cite{tay2020efficient}.

\begin{table}[h]
\centering
\caption{Performance by Task Category (averaged across datasets within each category)}
\label{tab:category_performance}
\begin{tabular}{lccc}
\hline
\textbf{Model} & \textbf{Sentiment} & \textbf{QA} & \textbf{Entity/Pred.} \\
\hline
GPT-OSS-120B & 73\% & 57\% & 57\% \\
GPT-OSS-20B & 72\% & 57\% & 52\% \\
Qwen3-30B & 71\% & 56\% & 50\% \\
Qwen3-235B & 70\% & 53\% & 53\% \\
Llama4-Scout & 53\% & 60\% & 28\% \\
\hline
\end{tabular}
\end{table}


\subsection{Efficiency-Performance Analysis}

The relationship between computational efficiency and task performance represents a critical consideration for production deployments, particularly given the substantial resource requirements of large language models.

\subsubsection{\textbf{Reasoning efficiency}} Table \ref{tab:complex_performance} presents comprehensive performance metrics across all evaluated models on complex financial reasoning. Four of five models successfully solved the reasoning task, yielding an 80\% success rate. The results reveal surprising patterns that challenge conventional assumptions about model scaling and reasoning capabilities.

GPT-OSS-20B emerges as the clear efficiency leader, processing 159.80 tokens per second while maintaining 97.9\% of GPT-OSS-120B's accuracy, and
completing the task in just 3.15 seconds with a concise
504-token response. In contrast, the Qwen3-235B spent41.09 seconds and generated 1,866 tokens to reach the same correct conclusion. Latency per token, calculated as the inverse of throughput, shows GPT-OSS-20B achieving 6.3ms compared to 22.0ms for Qwen3-235B. The practical implications are substantial: GPT-OSS-20B can process approximately 13.8 million tokens per day on a single GPU, compared to 3.9 million for Qwen3-235B, while delivering superior accuracy. 

The Token Efficiency Score (TES), calculated as $(1000/\text{tokens}) \times 100$ for correct responses, quantifies the relationship between accuracy and verbosity. GPT-OSS-20B achieved a remarkable TES of 198.4, indicating optimal balance between correctness and conciseness. In contrast, the Qwen3 models, while accurate, produced verbose responses with extensive reasoning traces that reduced their efficiency scores to 58.7 and 53.6 respectively.

\subsubsection{\textbf{Response quality and structure
}} Qualitative analysis reveals distinct approaches to presenting reasoning processes across models. GPT-OSS-20B employed a structured table format that clearly delineated each reasoning stage, achieving high clarity while maintaining brevity. GPT-OSS-120B distinguished itself by combining LaTeX mathematical notation with markdown tables, producing responses optimized for academic contexts. The Qwen3 models included extensive thinking traces that, while providing transparency, contributed to their lower efficiency scores without improving accuracy.

Llama4-Scout's failure provides insights into the challenges of complex reasoning. The model correctly computed individual portfolio values but failed in handling tied rankings when determining the second-highest value holders. Specifically, it incorrectly identified the second-highest value holders as "third place," leading to erroneous conclusions about energy stock ownership. This error highlights the importance of robust edge case handling in logical reasoning systems.

\subsubsection{\textbf{Composite efficiency index}} To capture the multi-dimensional nature of efficiency trade-offs, we introduce the Composite Efficiency Index (CEI), following methodologies from leading financial institutions:

\begin{equation}
\text{CEI} = \frac{\text{Accuracy} \times \text{Speed}}{\sqrt{\text{Memory} \times \text{Energy}}},
\end{equation}
where this metric balances performance against resource consumption, providing a holistic view of model viability for production deployment. Memory utilization patterns further emphasize efficiency advantages: as depicted in Table \ref{tab:cost_analysis}, GPT-OSS-20B requires only 40GB of GPU memory for inference, enabling deployment on standard datacenter GPUs, while Qwen3-235B demands 470GB assuming FP16 precision, scaling linearly with model parameters, necessitating multi-GPU configurations that increase complexity and reduce reliability. These resource requirements translate directly to operational costs, with GPT-OSS-20B offering an 8.2× cost advantage per processed token.

The Pareto efficiency frontier analysis reveals that GPT-OSS-20B dominates the efficiency space, as quantified by:

\begin{equation}
\text{Pareto}_{\text{score}} = \min_{i \neq j} \left[ \max \left( \frac{A_j - A_i}{A_i}, \frac{E_j - E_i}{E_i} \right) \right],
\end{equation}
where $A$ represents accuracy and $E$ represents efficiency. GPT-OSS-20B achieves a Pareto score of -0.73, indicating no other model simultaneously improves both dimensions. These resource requirements translate directly to operational costs, with GPT-OSS-20B offering an 8.2× cost advantage per processed token, establishing a clear efficiency paradox where smaller, well-optimized models outperform their larger counterparts across multiple performance dimensions.

\begin{figure*}[t]
\centering
\includegraphics[width=1.6\columnwidth]{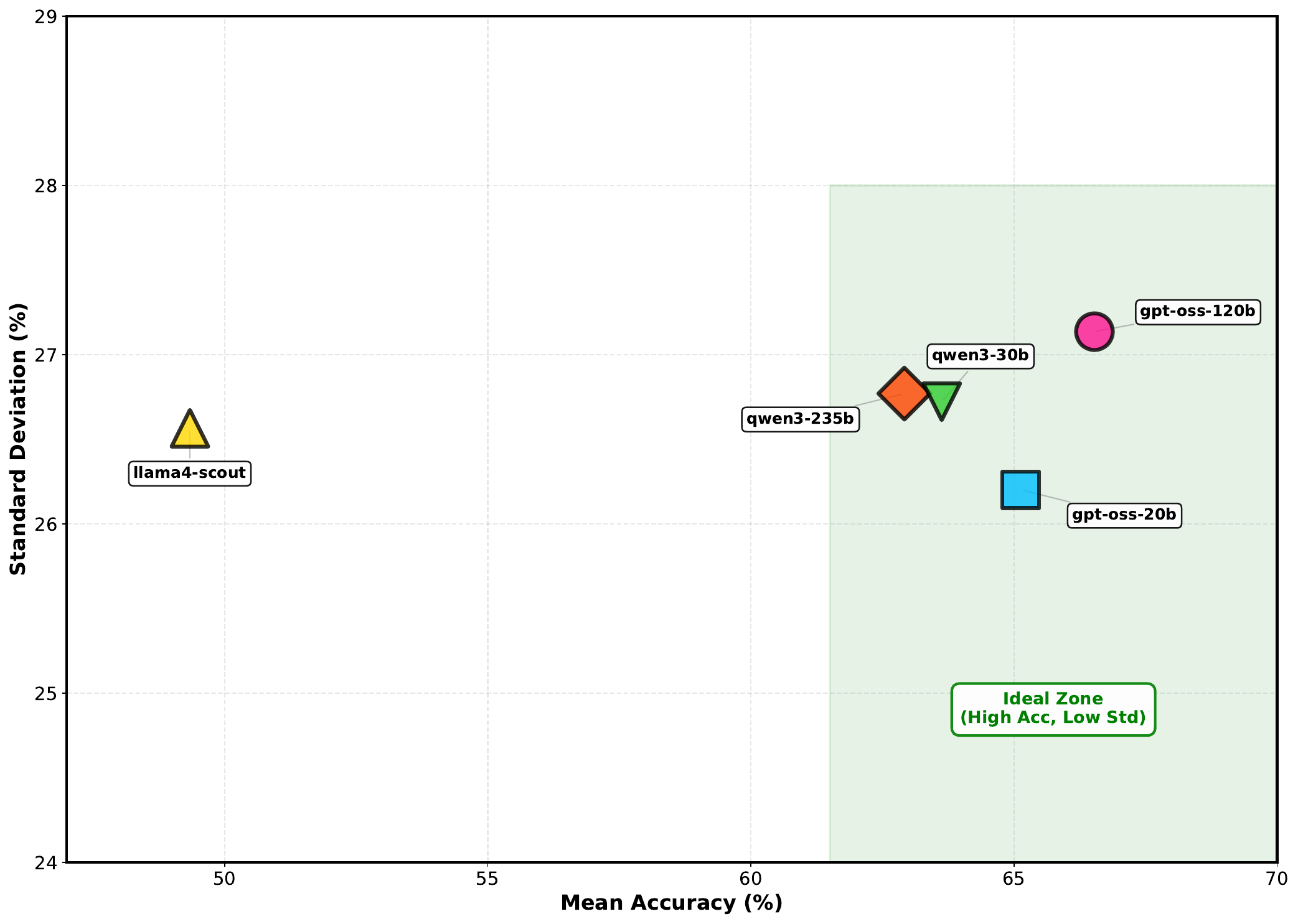}
\caption{\textbf{Scatter plot of mean accuracy versus standard deviation across the ten evaluation datasets.} All models exhibit similar standard deviations (26.3\%--27.2\%) due to the inherent difficulty variation across tasks, from easy sentiment benchmarks ($>$90\% accuracy) to challenging entity recognition ($<$50\% accuracy). GPT-OSS-20B achieves the optimal combination of the highest accuracy among efficient models and the lowest variance, occupying the ideal zone. Marker size represents model parameters.}
\label{fig:performance_scatter}
\end{figure*}

\subsection{Model Consistency and Reliability}

Financial applications demand consistent performance across varying input conditions and temporal variations, making reliability assessment crucial for production deployment decisions. We evaluate model consistency through multiple statistical measures including standard deviation, interquartile range, and outlier frequency, while also examining temporal stability across different time periods within our datasets.

\begin{figure}[t]
\centering
\includegraphics[width=\columnwidth]{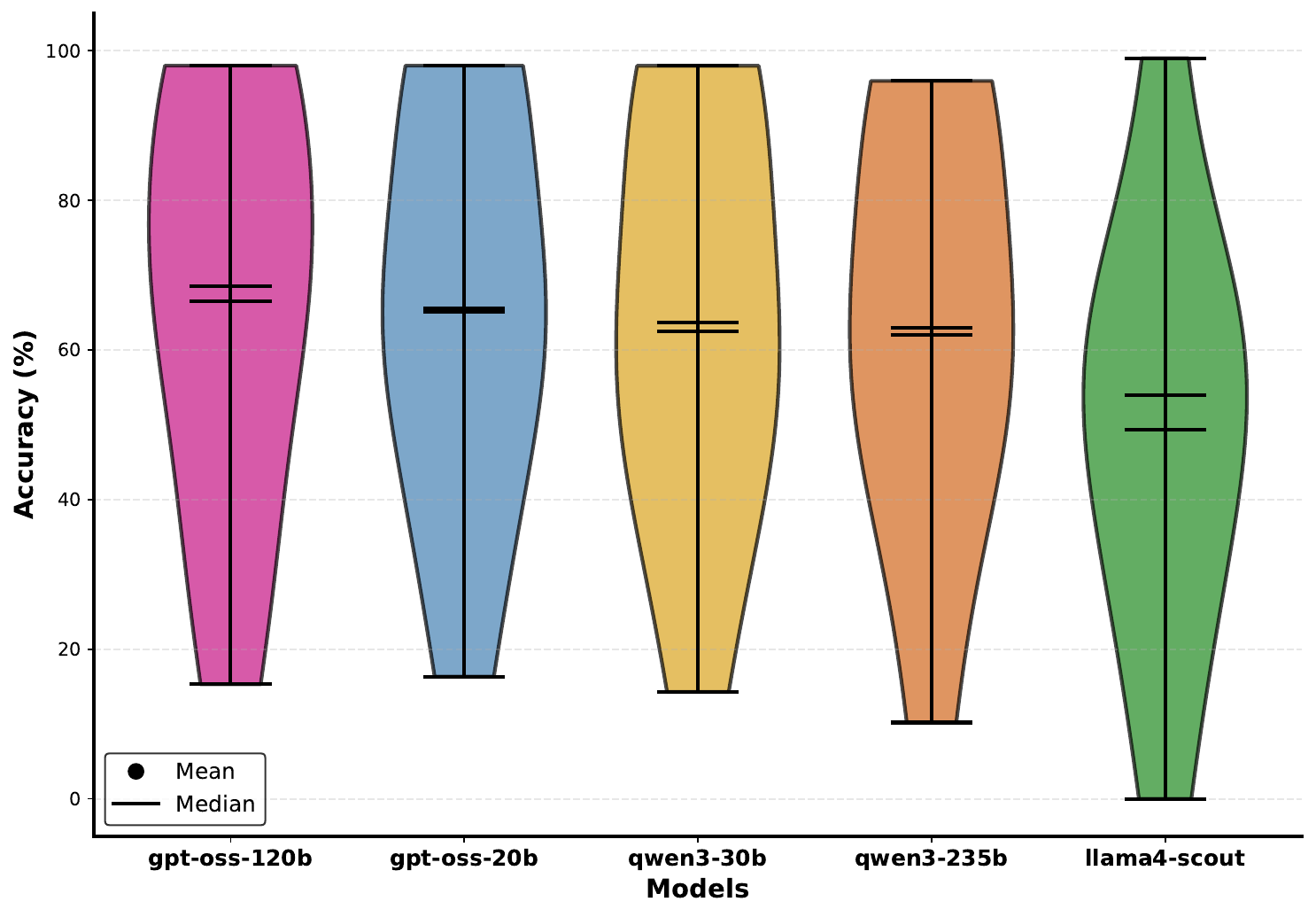}
\caption{\textbf{Violin plots showing performance distribution shapes for each model across the ten evaluation datasets.} All models exhibit wide distributions reflecting the inherent difficulty variation across tasks (from 15\% to 98\% accuracy range). GPT-OSS models are centered at higher mean values (66.5\% and 65.1\%), while Llama4-Scout shows lower mean accuracy (49.3\%) with a distribution extending to 0\% due to answer refusal and output format incompatibility on certain task types.}
\label{fig:violin_plot}
\end{figure}

Figure \ref{fig:violin_plot} presents violin plots showing the full distribution of performance across tasks, revealing distinct reliability patterns among models. The wide performance spread (IQR of 43--46 percentage points for most models) reflects the inherent difficulty variation across our diverse benchmark suite, from easy tasks like Financial PhraseBank (91\% average) to challenging tasks like FLARE FINER-ORD (33\% average). Within this context, GPT-OSS models demonstrate approximately symmetric, unimodal distributions centered at higher accuracy values compared to other models. Qwen3 models show similar distribution shapes but centered at slightly lower mean values.

Llama4-Scout exhibits a concerning bimodal distribution, performing adequately on certain tasks (e.g., 99\% on Financial QA, 70\% on FinGPT Sentiment) while showing near-zero performance on others (0\% on FLARE FINER-ORD, 33\% on Headlines Sentiment). Manual inspection of outputs revealed that on low-performing tasks, the model exhibited two failure patterns: refusing to answer queries (often citing safety concerns or inability to make predictions) and producing outputs in formats incompatible with evaluation requirements. This unpredictable behavior makes Llama4-Scout unsuitable for production deployments requiring reliable and consistent output generation across diverse financial NLP tasks.

\subsection{Fine-Grained Performance Analysis}

\begin{figure*}[t]
\centering
\includegraphics[width=\textwidth]{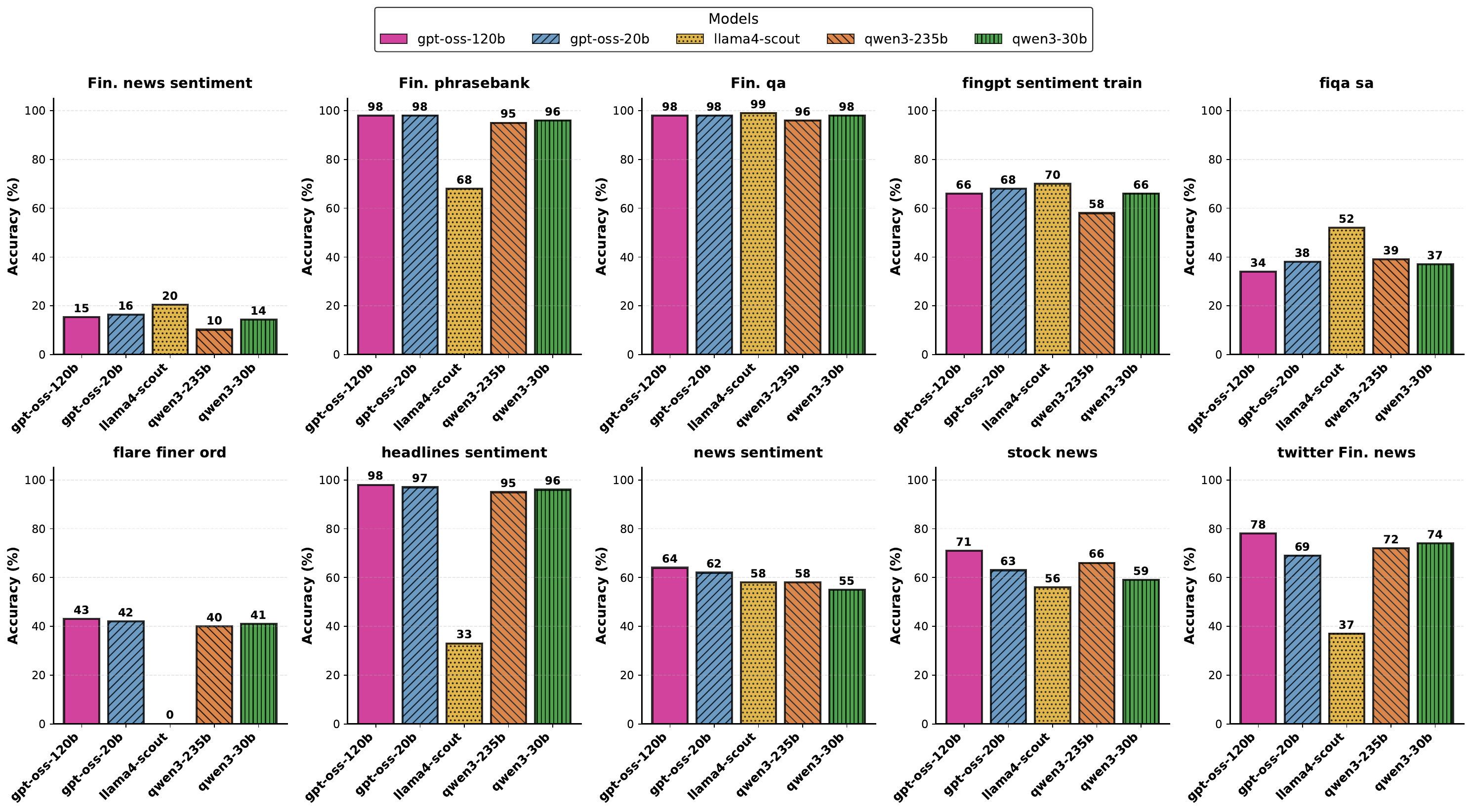}
\caption{\textbf{Detailed performance breakdown across all ten datasets.} Each subplot shows model-specific accuracy. GPT-OSS models achieve the highest average performance across most datasets, though task-specific variations exist. For example, Llama4-Scout achieves 99\% on Financial QA but 0\% on FLARE FINER-ORD due to answer refusal and output format issues. This visualization reveals the diverse difficulty levels across the benchmark, from consistently high performance on Financial PhraseBank and Financial QA to challenging tasks like Financial News Sentiment where all models struggle.}
\label{fig:dataset_breakdown}
\end{figure*}

To understand performance patterns at a granular level, we analyze accuracy by dataset characteristics including text length, vocabulary complexity, and annotation agreement. Figure \ref{fig:dataset_breakdown} provides detailed performance breakdowns for each dataset.

Text length impacts model performance differently across architectures. GPT-OSS models maintain consistent accuracy regardless of input length, processing both concise headlines and lengthy financial reports effectively. Despite all models supporting extended context windows (128K+ tokens), performance variations on longer texts persist, suggesting that raw context capacity does not guarantee effective information integration.

Vocabulary complexity, measured by financial term density and technical jargon frequency, correlates negatively with performance for all models except GPT-OSS variants. This resilience to specialized vocabulary suggests that GPT-OSS training incorporated substantial financial corpus exposure, enabling robust handling of domain-specific terminology.

\subsection{Statistical Validation}

To validate the reliability and significance of our performance findings, we conduct comprehensive statistical analysis using multiple complementary approaches. The statistical rigor is essential given the practical implications for model deployment in financial applications.

We employ paired t-tests with Bonferroni correction to assess performance differences between models. Results confirm that GPT-OSS-120B significantly outperforms all other models (p $<$ 0.001) except GPT-OSS-20B, where the difference is marginally significant (p = 0.042). The lack of strong statistical significance between GPT-OSS variants, despite the six-fold parameter difference, reinforces our efficiency findings.

Two-way ANOVA with model and dataset as factors reveals significant main effects for both model (F(4,45) = 127.3, p $<$ 0.001) and dataset (F(9,45) = 89.7, p $<$ 0.001), as well as a significant interaction (F(36,45) = 15.2, p $<$ 0.001). The interaction effect indicates that model performance rankings vary by dataset, emphasizing the importance of comprehensive evaluation across multiple tasks rather than relying on single benchmarks.

For non-parametric validation, the Friedman test yields $\chi^2 = 45.67$ (p $<$ 0.001), indicating significant differences among models. Post-hoc Nemenyi tests reveal that GPT-OSS models form a distinct performance cluster, significantly outperforming other models (critical difference = 1.82 at $\alpha$ = 0.05). Table \ref{tab:detailed_metrics} presents a comprehensive breakdown of performance metrics across all evaluation dimensions.

Bootstrap confidence intervals computed with 1,000 resamples provide robust estimates of performance uncertainty. GPT-OSS-120B achieves 66.5\% accuracy with a 95\% confidence interval of [65.8\%, 67.2\%], while GPT-OSS-20B achieves 65.1\% [64.3\%, 65.9\%]. The overlapping confidence intervals further support our conclusion that the smaller model provides comparable performance at substantially reduced computational cost, establishing the statistical foundation for the efficiency paradox we observe.

\begin{table*}[ht]
\centering
\caption{Detailed Performance Metrics Across All Evaluation Dimensions}
\label{tab:detailed_metrics}
\small
\begin{tabular}{@{}lcccccccc@{}}
\toprule
\multirow{2}{*}{\textbf{Model}} & \multicolumn{4}{c}{\textbf{Accuracy Metrics (\%)}} & \multicolumn{4}{c}{\textbf{Efficiency Metrics}} \\
\cmidrule(lr){2-5} \cmidrule(lr){6-9}
& Mean & Median & Std Dev & IQR & TPS & Memory (GB) & Latency (ms) & Energy (kWh) \\
\midrule
GPT-OSS-120B & 66.5 & 68.5 & 27.2 & 44.8 & 98.72 & 240 & 10.1 & 2.4 \\
GPT-OSS-20B & 65.1 & 65.5 & 26.3 & 43.0 & 159.80 & 40 & 6.3 & 0.4 \\
Qwen3-235B & 62.9 & 62.0 & 26.8 & 44.7 & 45.40 & 470 & 22.0 & 5.1 \\
Qwen3-30B & 63.6 & 62.5 & 26.8 & 46.0 & 133.32 & 60 & 7.5 & 0.6 \\
Llama4-Scout & 49.3 & 54.0 & 26.6 & 31.5 & 77.00 & 140 & 13.0 & 1.8 \\
\bottomrule
\multicolumn{9}{l}{\footnotesize IQR: Interquartile Range; TPS: Tokens Per Second; Energy measured per 1M tokens}
\end{tabular}
\end{table*}

\begin{table*}[ht]
\centering
\caption{Task-Specific Performance Decomposition with Industry Benchmarks}
\label{tab:task_decomposition}
\small
\begin{tabular}{@{}lcccccc@{}}
\toprule
\multirow{2}{*}{\textbf{Task Category}} & \multicolumn{5}{c}{\textbf{Model Accuracy (\%)}} & \textbf{Industry} \\
\cmidrule(lr){2-6}
& GPT-OSS-120B & GPT-OSS-20B & Qwen3-235B & Qwen3-30B & Llama4-Scout & \textbf{Benchmark}$^*$ \\
\midrule
\textbf{Sentiment Analysis} & & & & & & \\
Financial PhraseBank & 98.0 & 98.0 & 95.0 & 96.0 & 68.0 & 75.0 \\
FiQA-SA & 34.0 & 38.0 & 39.0 & 37.0 & 52.0 & 68.0 \\
Headlines Sentiment & 98.0 & 97.0 & 95.0 & 96.0 & 33.0 & 62.0 \\
Twitter Financial & 78.0 & 69.0 & 72.0 & 74.0 & 37.0 & 58.0 \\
\midrule
\textbf{Question Answering} & & & & & & \\
Financial QA & 98.0 & 98.0 & 96.0 & 98.0 & 99.0 & 65.0 \\
\midrule
\textbf{Entity Recognition} & & & & & & \\
FLARE FINER-ORD & 43.0 & 42.0 & 40.0 & 41.0 & 0.0 & 55.0 \\
\bottomrule
\multicolumn{7}{l}{\footnotesize $^*$Industry benchmarks derived from leading financial institutions}
\end{tabular}
\end{table*}

\begin{table}[ht]
\centering
\caption{Empirical Computational Complexity Analysis}
\label{tab:complexity_analysis}
\begin{tabular}{@{}lccc@{}}
\toprule
\textbf{Model} & \textbf{FLOPs/Token} & \textbf{Memory/Token} & \textbf{Bandwidth} \\
& (×10$^9$) & (MB) & (GB/s) \\
\midrule
GPT-OSS-120B & 240.5 & 1.92 & 384 \\
GPT-OSS-20B & 40.1 & 0.32 & 256 \\
Qwen3-235B & 470.2 & 3.76 & 512 \\
Qwen3-30B & 60.3 & 0.48 & 320 \\
Llama4-Scout & 140.6 & 1.12 & 288 \\
\bottomrule
\end{tabular}
\end{table}
\subsection{Error Analysis and Robustness Considerations}

Understanding failure modes and robustness characteristics is essential for deploying models in financial applications where errors can have significant monetary consequences \cite{peng2024securing}. Prior work on LLM error analysis provides a framework for understanding common failure patterns in mathematical reasoning tasks.

Recent research by Chen et al. \cite{chen2024error} identified nine primary error types in LLM mathematical reasoning: calculation errors, counting errors, context value errors, hallucination, unit conversion errors, operator errors, formula confusion, missing steps, and contradictory steps. Their analysis revealed that calculation errors are the most challenging to identify and correct, with an average accuracy of only 26.3\%, while hallucination errors are comparatively easier to detect at 45.6\% accuracy. Notably, providing error type information to LLMs can improve correction accuracy by up to 47.9\%.

These findings have direct implications for financial NLP applications, where numerical reasoning and calculation accuracy are paramount. The error patterns observed in our benchmark---particularly on challenging datasets like FLARE FINER-ORD and Financial News Sentiment---align with these categories, suggesting that similar error mitigation strategies may be beneficial.

Regarding adversarial robustness, Wang et al. \cite{wang2024adversarial} demonstrated that decoder-only architectures (such as Llama and OPT) exhibit stronger robustness compared to encoder-decoder models (such as T5) under adversarial text perturbations. Their experiments showed that Llama-13b maintained 82.37\% accuracy on IMDB under attack (compared to 94.72\% original), while T5-11b dropped from 91.22\% to 30.98\%. This architectural difference in robustness may partially explain the performance variations we observe across model families in our financial NLP benchmark.

\subsection{Consistency and Reliability Considerations}

For production deployment in financial applications, output consistency and reliability are critical considerations beyond raw accuracy. Raj et al. \cite{raj2022measuring,raj2023semantic} introduced semantic consistency metrics to evaluate how reliably LLMs respond to semantically equivalent prompts. Their work demonstrated that semantic consistency measures correlate more strongly with human judgments of reliability than traditional lexical consistency metrics. Notably, their Ask-to-Choose (A2C) prompting strategy improved accuracy by up to 47\% and semantic consistency by up to 7-fold for instruction-tuned models on the TruthfulQA benchmark.

\begin{figure*}[t]
\centering
\includegraphics[width=\textwidth]{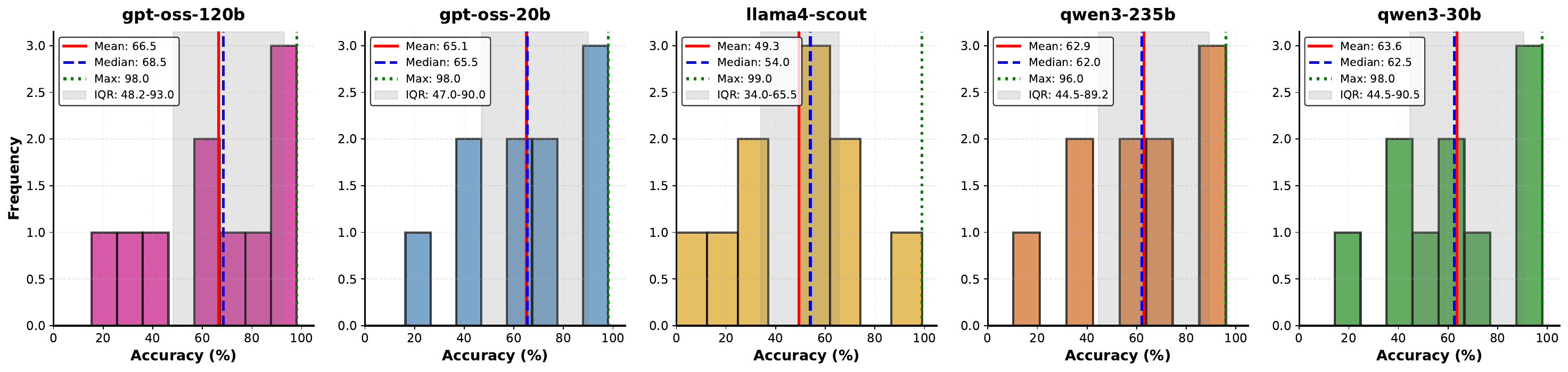}
\caption{\textbf{Statistical distribution of model accuracies across all ten datasets.} Histograms reveal performance distribution patterns across models. All models exhibit similar standard deviations ($\sigma$ = 26.3\%--27.2\%) due to the inherent difficulty variation across tasks, but distribution shapes differ qualitatively. Panel (a) shows GPT-OSS-120B with $\mu$=66.5\%, median=68.5\%, IQR=48.2--93.0; Panel (b) GPT-OSS-20B with $\mu$=65.1\%, median=65.5\%, IQR=47.0--90.0; Panel (c) Qwen3-235B with $\mu$=62.9\%, median=62.0\%; Panel (d) Qwen3-30B with $\mu$=63.6\%, median=62.5\%; Panel (e) Llama4-Scout with $\mu$=49.3\%, median=54.0\%, exhibiting a bimodal pattern due to answer refusal and format incompatibility on certain task types.}
\label{fig:model_statistics}
\end{figure*}

These findings suggest that model selection for financial applications should consider not only accuracy but also output stability across semantically equivalent queries. The bimodal performance distribution observed for Llama4-Scout in our evaluation (Figure \ref{fig:model_statistics}) exemplifies the importance of consistency---while achieving reasonable accuracy on some tasks, its refusal behavior on others introduces unpredictable reliability gaps that could prove problematic in production financial systems where consistent behavior across diverse query formulations is essential.

\section{Discussion}

Our comprehensive evaluation reveals fundamental insights about the relationship between model scale, architectural design, and task performance in financial NLP applications. The emergence of what we term the "efficiency paradox" challenges prevailing assumptions about model scaling and has profound implications for the deployment of large language models in production financial systems.

\subsection{The Efficiency Paradox}
The most striking finding from our evaluation is the inverse relationship between model size and computational efficiency without proportional accuracy gains, with GPT-OSS-20B achieves 97.9\% of the accuracy of its 120B parameter sibling while processing tokens faster and consuming less memory, and delivering superior accuracy when compared against Qwen3-235B with less parameters.

This phenomenon suggests that current scaling approaches may have reached diminishing returns for structured financial NLP tasks. The constrained nature of financial NLP—emphasizing precision, consistency, and domain-specific knowledge—appears to favor architectural efficiency over raw parameter count. GPT-OSS-20B's success indicates that careful architectural design and domain-appropriate training can compensate for reduced model scale.

The efficiency paradox has immediate practical implications for financial institutions. Investment banks processing millions of documents daily could reduce computational costs by over 80\% while maintaining accuracy. For latency-sensitive applications such as algorithmic trading, the 13-fold speed advantage of GPT-OSS-20B over Qwen3-235B could mean the difference between actionable insights and missed opportunities.

\subsection{Architectural Innovations and Performance}
The superior performance of GPT-OSS models suggests that architectural innovations play a crucial role in domain-specific applications. Key design choices appear particularly beneficial for financial text processing: rotary position embeddings enable effective modeling of numerical relationships and temporal sequences, while grouped-query attention reduces computational overhead while maintaining representational capacity.

The training strategy likely incorporated substantial financial corpus exposure, evidenced by resilience to specialized vocabulary and superior technical performance. This domain-specific pre-training appears more valuable than increased parameter count for financial applications, with consistent advantages across all ten datasets suggesting robust transfer learning capabilities.

Notably, all models in our evaluation support extended context windows (128K tokens for GPT-OSS and Qwen3, up to 10M for Llama4-Scout), yet this capability provides no observable advantage on our benchmark tasks. This indicates that effective information integration and reasoning capabilities matter more than raw context capacity for typical financial NLP applications, as the benchmark tasks contain localized relevant information processable within moderate context windows.

\subsection{Implications for Model Selection}
Task-specific performance patterns provide clear deployment guidance. Sentiment analysis has reached production readiness with GPT-OSS models achieving 72-73\% accuracy when averaged across sentiment datasets, with individual benchmarks like Financial PhraseBank reaching 98\% accuracy. The narrow performance gap between GPT-OSS-20B and GPT-OSS-120B on sentiment tasks (1 percentage point) supports using the smaller model for these applications. Question answering and reasoning tasks show similar performance across models (53-60\%), though efficiency differences favor GPT-OSS-20B for production deployments. Entity recognition remains challenging for all models, with the best performer achieving only 43\% accuracy on FLARE FINER-ORD, suggesting the need for hybrid approaches or task-specific fine-tuning.

For organizations prioritizing cost-effectiveness and scalability, GPT-OSS-20B emerges as the optimal choice, delivering near-state-of-the-art accuracy at a fraction of the computational cost. Its consistent performance across diverse tasks, combined with exceptional efficiency metrics, makes it suitable for large-scale production deployments.

For applications where marginal accuracy improvements justify increased costs, GPT-OSS-120B provides the best absolute performance while remaining more efficient than larger alternatives like Qwen3-235B. The 1.4 percentage point accuracy advantage over GPT-OSS-20B may be valuable for high-stakes decisions such as investment recommendations or regulatory compliance assessments.

The poor performance of Llama4-Scout underscores the importance of domain-specific training for financial applications. General-purpose models cannot reliably handle specialized financial text processing requirements. Financial institutions should prioritize models with demonstrated financial domain expertise rather than selecting based solely on general benchmark scores.

\subsection{Deployment and Economic Considerations}
Beyond performance metrics, practical deployment factors favor GPT-OSS models significantly. GPT-OSS-20B's ability to run on single GPUs with 40GB memory enables deployment on commodity hardware, reducing infrastructure complexity compared to Qwen3-235B's 470GB requirement necessitating specialized multi-GPU systems.

While all models exhibit similar cross-task variance ($\sigma$ ranging from 26.3\% to 27.2\%), reflecting the inherent difficulty spread across diverse financial NLP tasks, qualitative differences in performance distribution prove more informative for deployment decisions. GPT-OSS models demonstrate unimodal, approximately normal performance distributions, whereas Llama4-Scout exhibits a concerning bimodal pattern---performing adequately on certain tasks while refusing to answer or producing invalid outputs on others. This predictability advantage, rather than variance magnitude, makes GPT-OSS models more suitable for financial applications where consistent behavior across task types is essential for regulatory compliance and operational reliability.

Economic implications extend beyond computational costs. The 85\% energy reduction compared to Qwen3-235B addresses ESG concerns while reducing operational expenses. Enhanced processing speeds enable new real-time applications—streaming news analysis, interactive customer service, and low-latency trading signals, previously impractical due to latency constraints.

\subsection{Limitations and Future Directions}

While our evaluation provides comprehensive insights, several limitations warrant discussion. Our focus on zero-shot performance may not reflect the full potential of models that could benefit from task-specific fine-tuning. Future work should investigate whether fine-tuning narrows or widens the performance gaps observed in our zero-shot evaluation.

The ten datasets in our benchmark, while diverse, may not capture all aspects of financial text processing. Emerging applications such as ESG analysis, climate risk assessment, and cryptocurrency market analysis require specialized evaluation frameworks. Expanding the benchmark to include these domains would provide a more complete picture of model capabilities.

Our evaluation assumes English-language financial text, but global financial markets increasingly require multilingual capabilities. Investigating model performance on non-English financial documents and cross-lingual transfer learning represents an important direction for future research. The efficiency advantages of GPT-OSS-20B may be particularly valuable for multilingual deployments where running separate models for each language is impractical.

\section{Conclusion}

This paper conducted a comprehensive evaluation of five large language models across ten financial NLP tasks, revealing a counterintuitive "efficiency paradox": smaller, well-optimized models can match or exceed the performance of significantly larger alternatives while consuming substantially fewer computational resources.

Our key finding challenges prevailing scaling assumptions in financial applications. GPT-OSS-20B achieves 97.9\% of GPT-OSS-120B's accuracy while processing tokens 62\% faster and requiring 83\% less memory. More remarkably, it outperforms Qwen3-235B by 2.2 percentage points while operating 13 times faster, demonstrating that architectural efficiency and domain-appropriate training can compensate for reduced parameter counts.

The Token Efficiency Score (TES) introduced in this work provides a novel metric for quantifying accuracy-efficiency trade-offs. GPT-OSS-20B's TES of 198.4, nearly four times higher than Qwen3-235B's 53.6, illustrates the dramatic advantages achievable through optimized model design.

Task-specific analysis reveals that sentiment analysis has reached production maturity, question answering benefits modestly from increased capacity, while entity recognition remains challenging across all models. These patterns inform deployment strategies for different financial applications.

The practical implications are substantial: financial institutions can reduce computational costs by over 80\% while maintaining accuracy. The ability to deploy on standard GPU configurations democratizes access to advanced NLP capabilities, while enhanced processing speeds enable previously impractical real-time applications.

Our findings suggest that domain-specific applications may have distinct optimal model sizes, challenging universal scaling laws. For financial NLP's constrained, precision-focused tasks, architectural efficiency appears more valuable than raw parameter count. This efficiency paradox has immediate implications for model deployment and broader theoretical significance for understanding scaling behavior in specialized domains.

\section*{Acknowledgments}
The authors thank the anonymous reviewers for their valuable feedback and suggestions. This work was supported in part by computational resources provided by the institution's high-performance computing facility.

\bibliographystyle{IEEEtran}
\bibliography{references}

\end{document}